\documentclass[10pt,twocolumn,letterpaper]{article}

\usepackage{cvpr}              
\usepackage[accsupp]{axessibility} 
\usepackage{graphicx}
\usepackage{amsmath}
\usepackage{amssymb}
\usepackage{booktabs}
\usepackage{multirow}
\usepackage[numbers]{natbib}
\usepackage{color}

\usepackage{algorithm}
\usepackage{bm}
\usepackage{bbm}

\def\fullname{Visual-Dynamic Injection\xspace}
\def\abbrname{VDI\xspace}

\def\TransDe{\text{TransDecoder}}
\def\SeqModel{\text{SeqModel}}
\def\fc{\text{FC}}
\def\cossim{\text{cos}}
\def\convone{\text{Conv1D}}

\usepackage[pagebackref,breaklinks,colorlinks]{hyperref}

\usepackage[capitalize]{cleveref}
\crefname{section}{Sec.}{Secs.}
\Crefname{section}{Section}{Sections}
\Crefname{table}{Table}{Tables}
\crefname{table}{Tab.}{Tabs.}

\def\cvprPaperID{2714} 
\def\confName{CVPR}
\def\confYear{2023}

\begin{document}

\title{
Towards Generalisable Video Moment Retrieval: \\
Visual-Dynamic Injection to Image-Text Pre-Training
}
\author{
Dezhao Luo\textsuperscript{\rm 1},
Jiabo Huang\textsuperscript{\rm 1}, 
Shaogang Gong\textsuperscript{\rm 1}, 
Hailin Jin\textsuperscript{\rm 2}, 
and Yang Liu\textsuperscript{\rm 3}\thanks{Corresponding authors} 
\\
\small \textsuperscript{1}{Queen Mary University of London}\\
\tt\small \{dezhao.luo, jiabo.huang, s.gong\}@qmul.ac.uk\\
\small \textsuperscript{2}{Adobe Research}, \small \textsuperscript{3}{WICT, Peking University}\\
\tt\small hljin@adobe.com, \tt\small yangliu@pku.edu.cn
}


\maketitle

\begin{abstract}

The correlation between the vision and text is essential for video moment retrieval (VMR), however, existing methods heavily rely on separate pre-training feature extractors for visual and textual understanding. Without sufficient temporal boundary annotations, it is non-trivial to learn universal video-text alignments. In this work, we explore multi-modal correlations derived from large-scale image-text data to facilitate generalisable VMR.   
To address the limitations of image-text pre-training models on capturing the video changes, we propose a generic method, referred to as \fullname (\abbrname), to empower the model's understanding of video moments. 
Whilst existing VMR methods are focusing on 
building temporal-aware video features,
being aware of the text descriptions about the temporal changes is also critical
but originally overlooked in pre-training by matching static images with sentences.
Therefore,
we extract visual context and spatial dynamic information
from video frames
and explicitly enforce their alignments with
the phrases describing video changes (\eg verb).
By doing so,
the potentially relevant visual and motion patterns in videos are 
encoded in the corresponding text embeddings (injected)
so to enable more accurate video-text alignments.
We conduct extensive experiments on two VMR benchmark datasets (Charades-STA and ActivityNet-Captions)  and achieve state-of-the-art performances.  Especially, \abbrname yields notable advantages when being tested on the
out-of-distribution splits where the testing samples involve
novel scenes and vocabulary.
    
\end{abstract}
\begin{figure}[t]
  \centering
  \includegraphics[width=1\columnwidth]{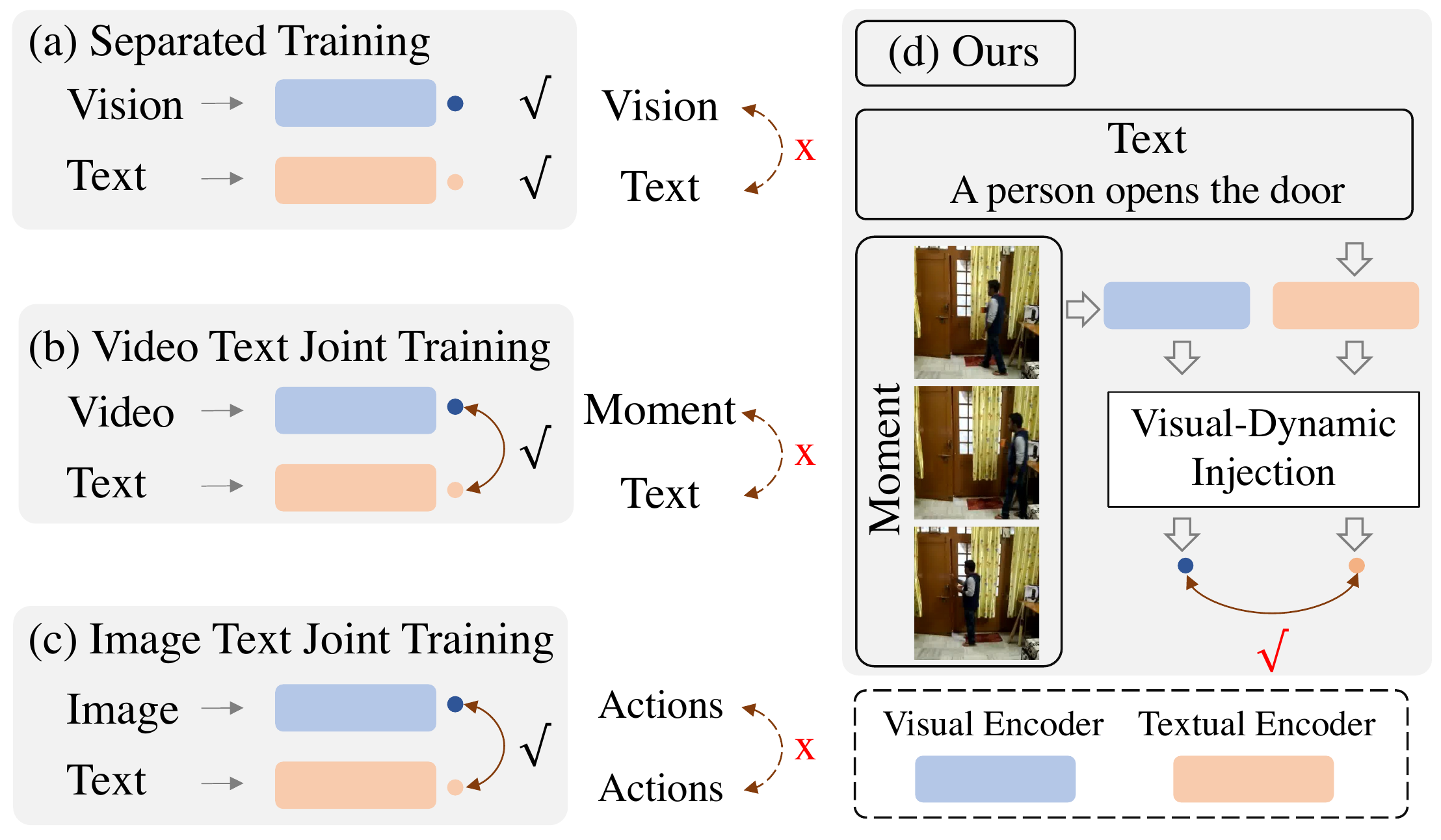}
  \caption{
   Contemporary methods lack moment-text correlations. Our
   method takes the advantage of image-text pre-trained models
   and learns moment-text correlations by visual-dynamic injection.} 
    
  \label{fig:intro}
\end{figure}
\section{Introduction}
\label{sec:intro}

Video moment retrieval (VMR) aims at
locating a video moment by its temporal boundary
in a long and untrimmed video
according to a natural language sentence~\cite{chen2018temporally,gao2017tall}.
It is a critical task
which has been extensively studied in 
a variety of real-world applications including 
human-computer interaction~\cite{chiyangwa2021natural}, 
and intelligent surveillance~\cite{dilawari2021natural}.
In practice, 
raw videos are usually unscripted and unstructured,
while the words being chosen 
for describing the same video moments
can be varied from person to person~\cite{sapir1927speech,zhou2021embracing}.
To be generalisable to different scenes,
VMR is fundamentally challenging
as it requires the comprehension of 
arbitrary complex visual and motion patterns in videos 
and an unbounded vocabulary with their intricate relationships.

For the fine-grained retrieval objective of VMR,
the precise segment-wise temporal boundary labels 
are intuitively harder to be collected than 
conventional image/video-level annotations.
In this case,
rather than training from scratch with 
a limited number of temporally labelled videos,
existing VMR solutions~\cite{ghosh2019excl,chen2018temporally,gao2017tall,2dtan} 
heavily rely on single-modal pre-training~\cite{vgg,devlin2018bert}
for visual and textual understanding (Fig.~\ref{fig:intro} (a)).
By doing so,
they focus on modelling the correlations between 
the pre-learned features of videos and sentences.
Nonetheless,
without sufficient training data,
it is non-trivial to derive universal video-text alignments
so to generalise to
novel scenes and vocabulary.

Separately,
the recent successes achieved by joint vision-language pre-training
in zero-shot learning~\cite{clip,align}
demonstrate the potential of adapting the multi-modal correlations
derived from large-scale visual-textual data
to facilitate generalisable VMR.
Whilst it is intuitive to adopt 
the video-text pre-learned features~\cite{miech2019howto100m,sun2019videobert,liu2021hit}
for moment retrieval (Fig.~\ref{fig:intro} (b)),
it has been shown that the models pre-trained with 
coarse-grained video-level labels
can not transfer well to localisation-based tasks like VMR
due to their unawareness of 
fine-grained alignments between text and frames or clips~\cite{cao2022locvtp}.
Such a misalignment problem is less likely to exist in pre-training by image-text matching.
However, 
image-based pre-training models~\cite{clip,align} are 
less sensitive to the changes in videos
and the words describing such dynamics in text~\cite{hendricks2021probing}.
This is inherent in matching sentences and images with static content
but is significant in understanding video actions and activities (Fig.~\ref{fig:intro} (c)).
It is suboptimal to directly apply image-text pre-learned features
on VMR.

In this work,
we propose a generic method for
exploiting large-scale image-text pre-training models
to benefit generalisable VMR
by the universal visual-textual correlations derived in pre-training,
dubbed as \fullname (\abbrname).
The key idea is to
explore the visual context and spatial dynamic information from videos
and inject that into text embeddings
to explicitly emphasise
the phrases describing video changes (\eg verb) in sentences (Fig.~\ref{fig:intro} (d)).
Such visual and dynamic information in text 
is critical for 
locating video moments composed of arbitrary evolving events
but unavailable or overlooked in image-text pre-training.
Specifically,
we consider it essential for VMR models to answer two questions: 
``what are the objects" and ``how do the objects change". 
The visual context information indicates the content in the frames,
\eg backgrounds (scenes), appearances of objects, poses of subjects, \etc.
Meanwhile,
the spatial dynamic is about the location changes of 
different salient entities in a video,
which potentially implies the development of their interactions.
\abbrname is a generic formulation,
which can be integrated into any existing VMR model.
The only refinement is to adapt the text encoder by 
visual-dynamic information injection
during training.
Hence,
no additional computation costs are introduced in inference.

Our contributions are three-folded:
\textbf{(1)}
To our best knowledge,
this is the first attempt on 
injecting visual and dynamic information
to image-text pre-training models
to enable generalisable VMR.
\textbf{(2)}
We propose a novel method for VMR called
\fullname (\abbrname).
The \abbrname method is a generic formulation
that can be integrated into existing VMR models
and benefits them from
the universal visual-textual alignments 
derived from large-scale image-text data.
\textbf{(3)}
The \abbrname achieves the state-of-the-art performances
on two standard VMR benchmark datasets.
More importantly,
it yields notable performance advantages
when being tested on the out-of-distribution splits
where the testing samples
involve novel scenes and vocabulary.
\abbrname's superior generalisation ability
demonstrates its potential for adapting image-text pre-training
to video understanding tasks 
requiring fine-grained visual-textual comprehensions.

\begin{figure*}[t]
  \centering
  \includegraphics[width=2\columnwidth]{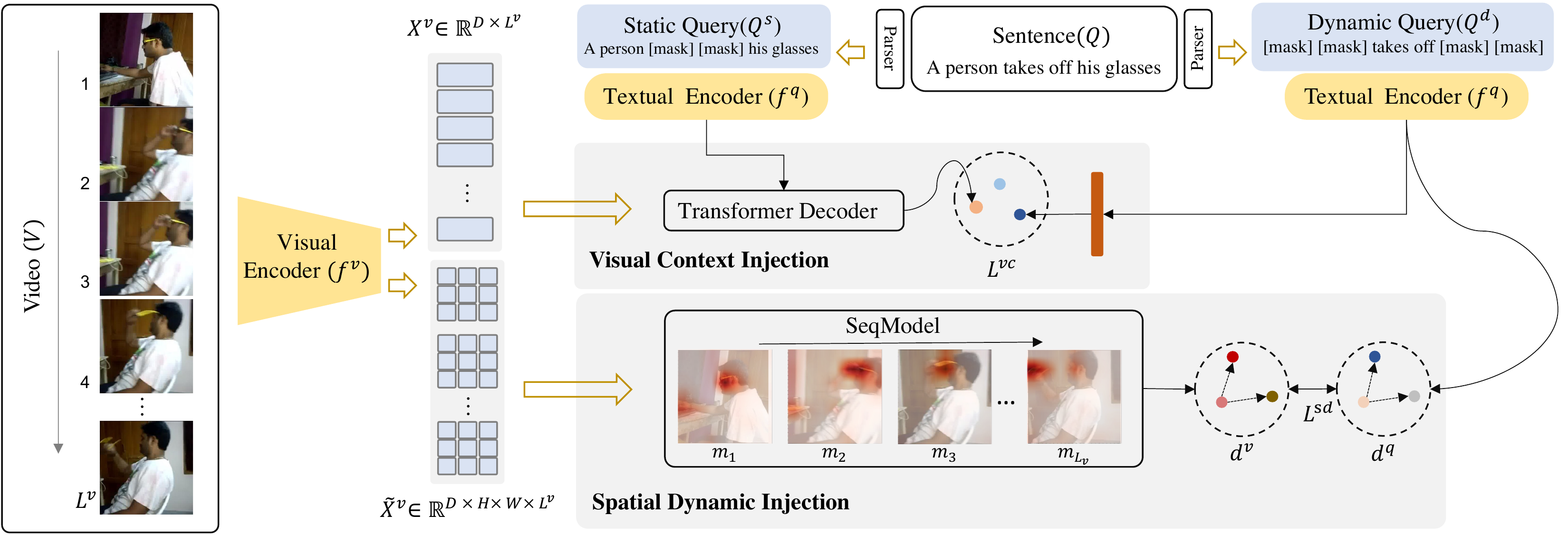}
  \caption{
   The framework of \abbrname, in which the video $V$ is fed into a visual encoder to generate image global features $X^v$ and image patch features $\tilde X^v$. The sentence $Q$ is parsed into static query $Q^s$ and dynamic query $Q^d$.  Visual Context Injection ($L^{vc}$) aligns the $Q^d$ with $Q^s$ guided visual context information. Spatial Dynamic Injection ($L^{sd}$) empowers the $Q^d$ with the awareness of the spatial dynamics. }
  \label{fig:method}
\end{figure*}

\section{Related Work}
\subsection{Video Moment Retrieval}

To tackle the VMR task and predict accurate moment boundaries, existing methods \cite{mgsl,vslnet,2dtan,mmn,huang2021cross,huang2022video} first generated visual features and textual features from pre-training encoders \cite{c3d,DistilBERT}, then they designed correlation models to align the two modalities.
\citet{ghosh2019excl}, \citet{cbln} and \citet{DRN} proposed to select the starting and ending frames by leveraging cross-modal interactions between text and video. 
\citet{he2019read} and  \citet{wang2019language} proposed reinforcement learning methods for VMR.
\citet{gao2017tall,2dtan,mmn} took a two-stage pipeline by generating proposals and ranking them relying on the similarity between proposal and query. 
\citet{dtsg,dcm,visa} focused on de-bias problems including the temporal location bias \cite{dcm,dtsg}, or the word-composition bias~\cite{visa}. 
 
Even though existing methods have demonstrated promising performance for VMR, we argue that models that take separate pre-training visual and textual feature extractors are suboptimal as they need to learn the alignment of the two modalities from scratch. 
It is demanding to learn from large-scale image-text datasets due to a lack of well-annotated moment-text datasets \cite{miech2019howto100m} resulting in poor generalisation \cite{cao2022locvtp}.

\subsection{Vision-Language Pre-Training}
Vision-language models have demonstrated
great potential in learning generic visual representations and allowing  transferring to a variety of
downstream tasks \cite{rao2022denseclip,kwon2022clipstyler}. Previously, \citet{mori1999image,fromedeep,weston2011wsabie} had explored the connection between images and words using paired text documents.
As more and more data accessible from the Internet, image-text pre-training models including CLIP \cite{clip}, ALIGN \cite{align}, ALBEF \cite{albef} and Florence~\cite{yuan2021florence} proposed to pre-train vision–language models
with a contrastive loss. Benefiting from large-scale web data (400M for CLIP, 1.8B for ALIGN and 900M for Florence), image-text pre-training methods can learn powerful visual representation as well as their correlations. 
Similar ideas can be seen in video-text pre-training \cite{cao2022locvtp} with a large-scale video-text dataset Howto100M \cite{miech2019howto100m}.

Even though the pre-training image-text models can capture the object appearance from visual embeddings and their corresponding description (\eg nouns) in the text embeddings \cite{subramanian-etal-2022-reclip}, we consider they are suboptimal to capture the change between frames as well as their corresponding descriptions  (\eg verbs)  in the text.

\subsection{Correlation Transfer Learning}

\citet{cao2022locvtp} explored the video-text pre-training to learn the alignment between moment and text. However, it is suboptimal with a lack of fine-grained well-annotated video-text alignment samples (50\% samples are not aligned in dataset Howto100M \cite{miech2019howto100m}). Meanwhile, image-text pre-training methods have shown promising generalisable ability with multiple downstream tasks, including image classification \cite{conde2021clip,long2022retrieval}, action recognition \cite{wang2021actionclip}, video retrieval~\cite{luo2022clip4clip}.
\citet{ju2022prompting} transferred CLIP models \cite{clip} to multiple video tasks by textual prompting and visual temporal modelling. \citet{wang2021actionclip} utilized CLIP models for video
recognition by traditional end-to-end fine-tuning. \citet{lin2022frozen} proposed to build temporal dependencies between images.  \citet{luo2022clip4clip,cheng2021improving} proposed to use correlation modelling ability from CLIP models and build their temporal relations by transformers\cite{vaswani2017attention} for video retrieval task.

Even though existing methods \cite{ju2022prompting,lin2022frozen,luo2022clip4clip,cheng2021improving} demonstrated promising results in transferring CLIP to video understanding task, a problem still remains: image-text pre-training models are less sensitive to actions\cite{hendricks2021probing}.  In this work, we explore multi-modal correlations derived from large-scale image-text data for generalisable VMR. To enable the action understanding for image-text pre-training models, we extract visual context information and spatial dynamic information and enable the text encoder to understand the entities in video frames and their movements. 

\section{VMR by Image-Text Pre-Training}
Given an untrimmed video $V=\{\bm{I}_i\}_{i=1}^{L^v}$ 
composed of $L^v$ frames
and a natural language sentence $Q$,
the objective of video moment retrieval
is to predict when the target video moment
starts $t^s \in [1, L^v]$ and ends $t^e \in [t^s, L^v]$ in the video $V$.
For generalisable VMR,
not only what happens in $V$ and what is described in $Q$ 
but also their alignments are supposed to be modelled.
This is intrinsically challenging
considering the free-form nature of both the unscripted videos
and natural language sentences
as well as their complex correlations.

To retrieve a target video moment by a text sentence,
the untrimmed video $V$ and query sentence $Q$ 
are first fed into a pre-training visual and a textual encoder in respective
to obtain the video $X^v = \{\bm{x}_i^v\}_{i=1}^{L^v} = \{f^v(\bm{I}_i)\}_{i=1}^{L^v} \in \mathbb{R}^{D\times L^v}$
and sentence $\bm{x}^q=f^q(Q) \in \mathbb{R}^D$ features.
Both the features will then be taken as the inputs to 
a video moment retrieval model $f^g(X^v, \bm{x}^q)$
to predict the temporal boundary of the target moment 
$(\tilde{t}^s, \tilde{t}^e)$.
As the visual and textual encoders are pre-trained
on image-text data,
they are unaware of temporal changes in videos
or the words/phrases describing them in the text.
Therefore,
we additionally adapt
the two encoders (Fig.~\ref{fig:method})
to model the visual context and spatial dynamic 
in both modalities.
Together with the pre-learned visual-textual correlations
which are prone to be universal,
the video change-sensitive features
will enable $f^g$
to predict accurate temporal boundary
even for video moments filmed in novel scenes or described by unseen vocabulary.

\subsection{Visual-Dynamic Injection}
We start with using a language parsing tool\footnote{spaCy: https://spacy.io/}
to extract all the noun chunks (a noun plus the words describing the noun)
in the query sentence $Q$,
which are supposed to describe the entities of interest
in the target video moment.
We mask all the other words in the sentence $Q$
and consider such a masked query 
can be matched with the corresponding video content
without knowing the temporal dependencies among frames.
In this case,
we call the masked sentence about the static content in videos
as \textit{static query} $Q^s$.
In contrast,
we construct another \textit{dynamic query} $Q^d$
with all the noun chunks in $Q$ being masked,
which is critical for understanding complex video moments
composed of arbitrary evolving events
but missing in the image-text pre-training.
Whilst the video change information can be captured
by additional sequence analysis in the video moment retrieval model $f^g$,
it is impractical for $f^g$ 
to understand the phrases describing video changes in the text
which are originally overlooked in $\bm{x}^q$.
Therefore,
we model video changes with visual context injection and spatial dynamic injection 
and enforce the text encoder to match the dynamic queries $Q^d$ with them.
By doing so,
the adapted text encoder
is able to yield visual-dynamic sensitive representations for query sentences
to ensure more accurate VMR
by both visual and dynamic matching. 

\paragraph{Visual Context Injection.}
The visual context we discover in the videos
is about how the frames display the entities related to the video changes.
Such visual information is likely to encoding
the presence of scenes (\eg outdoor or indoor),
the status of entities (\eg boiling or cold water) and \etc.
It is important for recognising and locating
video moments which involve specific objects and scenes.
To that end,
we apply a transformer decoder~\cite{vaswani2017attention}
and encode the visual context information
into the static query $Q^s$:
\begin{equation}
\begin{gathered}
\bm{x}^{qs} = f^q(Q^s) \in \mathbb{R}^{D}, \\
\tilde{x}^{qs} = \TransDe(\bm{x}^{qs}, X^v, X^v) \in \mathbb{R}^{D}. 
\label{eq:vis_dynamic}
\end{gathered}
\end{equation}
In Eq.~\eqref{eq:vis_dynamic},
$\bm{x}^{qs}$ is the $D$-dimensional textual feature of the static query $Q^s$
obtained by the text encoder $f^q$,
$X^v$ is the video frame features,
and $\tilde{x}^{qs}$
is the visual context-aware feature of $Q^s$
computed by a transformer decoder 
whose three inputs correspond to the query, key and value,
respectively.
To inject such visual context information to the text encoder with a focus on the words describing the  changes in videos,
we compute the dynamic query feature $Q^d$
by the text encoder
and encourage its consistency with $\tilde{\bm{x}}^{qs}$:
\begin{equation}
\begin{gathered}
\bm{x}^{qd} = f^q(Q^d) \in \mathbb{R}^{D} \\
\mathcal{L}^{vc}(V, Q) = \lVert \fc(\bm{x}^{qd}) - \tilde{\bm{x}}^{qs} \rVert^2_2,
\label{eq:loss_vc}
\end{gathered}
\end{equation}
where $\fc(\cdot)$ is a linear projection layer. 
The rationale behind this design
is to probe the video frames
by the static query
in order to
select the visual context engaging
the entities potentially existing in the target moments.
By doing so,
the text encoder is updated to
align the dynamic query with 
its visual context
and avoid distractions from irrelevant video content.

\paragraph{Spatial Dynamic Injection.}
Besides the visual context
demonstrating the entity of interests in frames,
another crucial information for video change
is about how the spatial locations
of different entities change in time order.
However,
the motion patterns encoding such dynamic information
is hidden in the complex visual patterns in videos.
It is non-trivial to discover and use them to 
raise the text encoder's attention
on the corresponding descriptions.
Therefore,
we propose to extract the location changes of salient entities in videos
and explicitly inject such spatial dynamics
into the text encoder.
To that end,
we first obtain the per-frame spatial features 
$\tilde{X}^{v}_i\in \mathbb{R}^{D\times H\times W}\ \forall\ i\in [1, L^v]$
as the last feature maps produced by the convolutional neural networks%
~\cite{lecun1989handwritten,Simonyan15,he2016deep}
or the patch-wise features in Visual Transformers~\cite{dosovitskiy2020vit}.
The $H$ and $W$ denote the height and width resolutions of concerns.
We then adopt a transformer-like formulation
to compute a heatmap for each video frame:
\begin{equation}
\begin{gathered}
M_i = \fc(\bm{x}^{v}_i)\cdot\fc(\tilde{X}^v_i) / \sqrt{D} \in \mathbb{R}^{H\times W}.
\end{gathered}
\end{equation}
The frame-wise heatmap is computed by
the correlations between every spatial feature
and the global image representation.
Therefore,
the salient entities whose visual information
is encoded in the image feature will result in
corresponding salient regions in the heatmap.
We then flatten the heatmap and feed it into a linear projection layer
to compute a $D$-dimensional vector
as the holistic representation of the spatial feature for each frame.
The spatial dynamics in the video can be given
by any sequence analysis model:
\begin{equation}
\begin{gathered}
\bm{m}_i = \fc(\text{Flatten}(M_i)) \in \mathbb{R}^{D} \\
\tilde{\bm{m}} = \SeqModel(\{\bm{m}_i\}_{i=1}^{L^v}) \in \mathbb{R}^{D}.
\end{gathered}
\end{equation}
As the visual information is deprecated in the spatial features,
we cannot probe them by the static query $Q^s$. Hence,
we choose a transformer encoder~\cite{vaswani2017attention} to build their dependencies and take the averaged outputs
as the spatial dynamic feature $\tilde{\bm{m}}$ of the video.

To inject such dynamic information
into the text encoder,
we then enforce consistent correlations 
between the spatial dynamic features of different videos ($V$ and $V'$)
and the corresponding descriptions in text ($Q^d$ and $Q^{d'}$) by
\begin{equation}
\begin{gathered}
\epsilon^v = \cossim(\tilde{\bm{m}}, \tilde{\bm{m}}')\quad
\epsilon^q = \cossim(\fc(\bm{x}^{qd}), \fc(\bm{x}^{qd'})) \\
\mathcal{L}^{sd}(V, Q, V', Q') = (\epsilon^v - \epsilon^q)^2.
\end{gathered}
\label{eq:loss_sd}
\end{equation}
In Eq.~\eqref{eq:loss_sd},
the notation $\epsilon^v$ stands for
the cosine similarity between 
the spatial dynamic features of two videos
while $\epsilon^q$ is that of 
the two corresponding dynamic queries $Q^d$ and $Q^{d'}$.
In contrast to the visual context injection,
since the spatial features used here lost all the visual cues in videos,
we optimise their correlations consistency with that of the dynamic queries 
rather than directly pushing them closer to the matched textual features.
By doing so,
different sentences
matched with the videos sharing similar motion patterns
will be encouraged to focus on 
the common descriptions of such dynamic information in the text.

\paragraph{Video Moment Retrieval.}

With 
the visual and textual features pre-learned from large-scale image-text datasets
as well as
our adaptation of textual features to be aware of
temporal changes in videos,
our \abbrname model is ready to benefit existing VMR models.
Here, we take the state-of-the-art Mutual Matching Network (MMN) as an example.
Specifically,
given the frame-wise video features $X^v$ and the sentence feature $\bm{x}^q$,
we first enumerate all the start-end frame pairs
to generate $L^m=L^v\times L^v$ video segments
as the candidates of the target moment.
We then construct a 2D feature map  $X^m=\text{Conv2D}(\{\bm{x}^m_{i,j}\}_{i,j=1}^{L^v})\in\mathbb{R}^{D\times L^v\times L^v}$
where $\bm{x}^m_{i,j}$
is the feature of the segment starting from the $i$-th frame
and ending at the $j$-th frame.
After that,
both the features of video segments $X^m$ and query sentences $\bm{x}^q$
will be linearly projected into a common space and
their alignments are then measured by cosine similarities $\cossim(\cdot,\cdot)$:
\begin{equation}
\tilde{Y}^\text{iou}=\cossim(\fc(\bm{x}^q), \convone(X^m)) \in \mathbb{R}^{L^v\times L^v}.
\label{eq:iou_scores}
\end{equation}
The predicted alignment scores $\tilde{Y}^\text{iou}$ 
between every segment and the query sentence
will be supervised by the temporal IoU between it 
and the manually labelled temporal boundary $(t^s, t^e)$
of the target moment:
\begin{equation}
\begin{gathered}
y^\text{iou}_{i,j} = \text{IoU}((t^s, t^e), (i, j)), \\
Y^\text{iou} = \{y^\text{iou}_{i,j}\}_{i,j=1}^{L^v} \in \mathbb{R}^{L^v\times L^v},\\
\mathcal{L}^{iou}(V, Q) = \text{BCE}(Y^\text{iou}, \tilde{Y}^\text{iou}).
\end{gathered}
\label{eq:loss_iou}
\end{equation}

Besides learning to locate video moments
by aligning positive segment-text pairs,
we follow MMN to conduct 
mutual-modal
contrastive learning
among negative sample pairs.
In particular,
for a moment $\bm{x}^{m}_{t^s, t^e}$
in video $V$ and its description $\bm{x}^{q}$,
we construct a negative video set $\mathcal{X}^{m-}$
and a negative sentence set $\mathcal{X}^{q-}$.
We then map the segments and queries features
to another shared feature space by linear projections
and conduct contrastive learning by:
\begin{equation}
\begin{gathered}
\tilde{X}^m = \{\tilde{\bm{x}}^m_{i,j}\}_{i,j=1}^{L^v} = \convone(X^m) \in \mathbb{R}^{D\times L^v \times L^v}, \\
\tilde{\bm{x}}^q = \fc(\bm{x}^q) \in \mathbb{R}^{D}, \\
p^{m} =
\frac{
\exp(\cossim(\tilde{\bm{x}}^{m}_{t^s, t^e}, \tilde{\bm{x}}^{q}) / \tau)}{
\sum_{\bm{x} \in \{\bm{x}^{m}_{t^s, t^e}\}\cup\mathcal{X}^{m-}}\exp(\cossim(\tilde{\bm{x}}, \tilde{\bm{x}}^{q}) /\tau)
} \\
p^q =
\frac{
\exp(\cossim(\tilde{\bm{x}}^{m}_{t^s, t^e}, \tilde{\bm{x}}^{q}) / \tau)}{
\sum_{\bm{x} \in \{\bm{x}^{q}\}\cup\mathcal{X}^{q-}}\exp(\cossim(\tilde{\bm{x}}^{m}_{t^e, t^e}, \tilde{\bm{x}}) /\tau)
} \\
\mathcal{L}^{cl}(V, Q) = -\log p^m - \log p^q.
\end{gathered}
\label{eq:loss_cl}
\end{equation}
In Eq.~\eqref{eq:loss_cl},
the tilde on top of all the segments and queries features 
$\{\bm{x}^m_{t^s, t^e}, \bm{x}^q, \bm{x}\}$
denotes their linear projected counterparts.
The variables $p^m$ and $p^q$
measure how likely the model identifies 
the target moment $\bm{x}^m_{t^s, t^e}$ and 
the query sentence $\bm{x}^q$
according to each other from the respective negative sets.
In this multi-modal common space,
we can compute another alignment scores between 
every candidate segment and the query sentence:
\begin{equation}
\tilde{Y}^{cl} = \cos(\tilde{\bm{x}}^q, \tilde{X}^m) \in \mathbb{R}^{L^v, L^v}.
\end{equation}
The two video-text alignment predictions
will then be fused by the hadamard product
and the temporal boundary predicted for the target moment
can then be computed in a maximum likelihood manner:
\begin{equation}
\begin{gathered}
\tilde{Y} = \tilde{Y}^{iou}\odot\tilde{Y}^{cl} \\
\tilde{t}^s = \arg\max(\text{cmax}(\tilde{Y})),\quad
\tilde{t}^e = \arg\max(\tilde{\bm{y}}_{t^s}).
\end{gathered}
\label{eq:predict}
\end{equation}
By learning from $\mathcal{L}^{iou}$ and $\mathcal{L}^{cl}$ jointly,
the model is trained to be aware of both the matched and unmatched
video-text information.
\begin{algorithm}[ht]
\caption{\fullname (\abbrname)}\label{alg}
\textbf{Input:} 
  Untrimmed videos $V$,
  Query sentences $Q$,
  Temporal boundary labels $(t^s, t^e)$,
  A visual $f^v$ and a textual encoder $f^q$ from image-text pre-training. \\
\textbf{Output:}
  An updated video moment retrieval model. \\
Generates the static $Q^s$ and dynamic $Q^d$ query sentences; \\
Computes the features of $Q^s$ and $Q^d$ by $f^q$; \\
Computes the features of videos $V$ by $f^v$; \\
Computes the visual context $\mathcal{L}^{vc}$ (Eq.~\eqref{eq:loss_vc}) and spatial dynamic  $\mathcal{L}^{sd}$ (Eq.~\eqref{eq:loss_sd})  losses; \\
Adapts the textual encoder $f^q$ by minimising $\mathcal{L}^{vc}$ and $\mathcal{L}^{sd}$; \\
Computes the features of the query $Q$ by $f^q$; \\
Feeds the features of video $V$ and query $Q$ to $f^g$; \\
Computes the losses $\mathcal{L}^{iou}$ (Eq.~\eqref{eq:loss_iou}) and $\mathcal{L}^{cl}$ (Eq.~\eqref{eq:loss_cl}); \\
Optimises the VMR model $f^g$ by minimising $\mathcal{L}^{iou}$ and $\mathcal{L}^{cl}$.
\end{algorithm}

\subsection{Model Training}
The \abbrname model is trained
in a conventional batch-wise manner.
A mini-batch is composed of $n$ randomly sampled
video-text pair $(V, Q)$ as well as 
the temporal boundary labels $(t^s, t^e)$
of the target moments.
The overall loss functions are computed by:
\begin{equation}
\begin{aligned}
\mathcal{L} = \frac{1}{n}\sum_{i=1}^n (
&\lambda^{iou}\mathcal{L}^{iou}(V_i, Q_i) +
\lambda^{cl} \mathcal{L}^{cl}(V_i, Q_i) + \\
&\lambda^{vc} \mathcal{L}^{vc}(V_i, Q_i) + \\
&\lambda^{sd} \frac{1}{n}\sum_{j=1}^n \mathcal{L}^{sd}(V_i, Q_i, V_j, Q_j)).
\end{aligned}
\label{eq:loss_all}
\end{equation}
The training objective function $\mathcal{L}$ in Eq.~\eqref{eq:loss_all}
is then be used to optimise both the VMR model $f^g$
and the text encoder $f^q$ from the pre-training model
by any stochastic gradient descent algorithms. 
The overall training process of the \abbrname model 
is summarised in Alg.~\ref{alg}.

\begin{table*}
 \renewcommand\arraystretch{1.05}

  \setlength{\tabcolsep}{0.35cm}
  \centering
  \begin{tabular}{l|c|c|ccc|ccc}
    \hline
     \multirow{2}{*}{Method}&\multirow{2}{*}{Year}&\multirow{2}{*}{Pre-train} &\multicolumn{3}{c|}{Charades-STA}&\multicolumn{3}{c}{ActivityNet-Captions} \\
     &&&IoU=0.5&IoU=0.7 & mIoU & IoU=0.5&IoU=0.7 & mIoU \\
    \hline
    WSSL \cite{duan2018weakly} & 2018&\multirow{7}{*}{\shortstack{Video\&Text \\ Separated}}& 2.79& 0.73&7.92 & 3.09&1.13&7.10 \\
        TMN \cite{tmn2018} &2018&& 9.43 & 4.96 & 11.23 & 9.93 & 5.12 & 11.38 \\
    TSP-PRL \cite{tsp}&2020&& 14.83 & 2.61 &14.03 & 18.05 & 3.15 & 14.34 \\

    2D-TAN \cite{2dtan}  & 2020&&29.36 &13.12 &28.47 & 23.86 & 10.37 & 28.88 \\
    LGI \cite{lgi2020} & 2020 && 26.48 & 12.47 & 27.62& 23.10 & 9.03 & 26.95 \\
      VSLNet \cite{vslnet} & 2020& &25.60 &10.07 & 30.21& 21.68 & 9.94 &29.58 \\
    VISA \cite{visa} & 2022&& 42.35 & 20.88 & 40.18& 30.14& 15.90 & \textbf{35.13} \\
    \hline
    MMN \cite{mmn}& 2022 &\multirow{2}{*}{\shortstack{Image-Text\\Joint}} & 43.85 & 24.17 & 39.50 &31.05&15.48&33.16 \\
    \abbrname(Ours)& 2023 &&\textbf{46.47}  &\textbf{28.63}& \textbf{41.60} & \textbf{32.35} & \textbf{16.02} & 34.32 \\
        \hline

  \end{tabular}
  \caption{ Novel-word testing comparison between our method with other state-of-the-art methods on Charades-STA \cite{gao2017tall} and ActivityNet-Captions \cite{krishna2017dense}. 
  The ``Pre-train'' column indicates the types of pre-trained models adopted. 
  }
  \label{table:novel word}
\end{table*}

\section{Experiments}
 To evaluate the importance of generalisable correlations between the visual and textual space, we conduct experiments on video moment retrieval (VMR) and compare with the SOTAs on both out-of-distribution (OOD) and independent and identically distributed (IID) data splits.  
In this section, we first explain the implementation details and then report our results in the comparison with recent methods. Finally, we evaluate the effectiveness of each component in our methods.

\subsection{Experimental Settings}
\subsubsection{Dataset}

\noindent \textbf{Charades-STA \cite{gao2017tall}} is a benchmark dataset for VMR, which is built upon the Charades 
dataset \cite{sigurdsson2016hollywood}. The Charades dataset is collected for video action
recognition and video captioning. \citet{gao2017tall} adapt the
Charades dataset to VMR by collecting
the query annotations. The Charades-STA dataset contains
6670 videos and involves 16124 queries, where
12404 pairs are used for training and 3720 for testing. The average
duration of the videos is 30.59 seconds and each
video contains 2.41 annotated moments, and the moment has an average duration of 8.09 seconds. 

\noindent \textbf{ActivityNet-Captions} \cite{krishna2017dense} is collected for the
dense video captioning task from ActivityNet \cite{caba2015activitynet} where the videos are associated with 200 activity classes,
and the content is more diverse compared to Charades-STA. The ActivityNet-Captions dataset consists of 19811 videos with 71957
queries.   The average duration of the videos is around 117.75 seconds and each video contains 3.63 annotated moments, and
the moment has an average duration of 37.14 seconds. 
The public split of the  ActivityNet-Captions dataset  contains a training set and two validation
sets val\_1 and val\_2, including 10009,
4917, 4885 videos separately.

 
\subsubsection{Evaluation Metrics}

We adopt “R@n, IoU = µ” and “mIoU”
as the evaluation metrics, where
“R@n, IoU = µ” denotes the percentage of language queries having at least one result whose intersection over union (IoU) with ground truth is larger
than µ in top-n retrieved moments. “mIoU” is the
average IoU over all testing samples. 
We report the results as n $\in \{1, 5\}$
with µ $\in \{0.5, 0.7\}$ for fair IID split comparison following MMN~\cite{mmn}  and  n$ \in \{1\}$
with  µ$ \in \{0.5, 0.7\}$ and mIoU for fair OOD split comparison with \cite{visa}.

\subsubsection{Implementation Details}

We experiment with the MMN \cite{mmn} as the VMR framework to evaluate our method. Specifically, we apply the pre-training visual extractor 
of the CLIP (ViT-B/32) \cite{clip} as the backbone, and directly
feed to the VMR framework to generate proposals. Our VMR framework is similar to MMN, where we only delete the linear layer in the pooling module  to maintain the feature structure of CLIP.
During training, we only update the parameters of the text encoder to empower the understanding of video change, no additional computation cost is introduced in inference.

We use AdamW \cite{loshchilov2017decoupled} optimizer with a
learning rate of $1 \times 10^{-4}$
and a batch size 48 for Charades-STA,
 a learning rate of  $8 \times 10^{-4}$
and a batch size 48 for ActivityNet-Captions.
Following MMN, we early stop the training when we observe the performance on the validation set
start to drop. The learning rate of text encoder is always
1/10 of the main model.  $\lambda_{vc}$ and  $\lambda_{sd}$ is set to 0.5 and 0.01.
\subsection{Comparison with the SOTAs}
In this section, we compare the results of our methods under the VMR task with the baseline MMN \cite{mmn} and existing SOTAs. To evaluate the importance of generalisability between the vision and text, we report the results under OOD testing and IID testing.

\subsubsection{Novel-Word OOD Testing}

To validate the generalisation ability of our method to capture unseen words and scenes, we conduct experiments on novel-word OOD testing. Specifically, the novel-word OOD split is recently released by \citet{visa} where the testing split contains novel words which are not seen in the training split, and the corresponding scenes are not seen as well. We follow the settings in  \citet{visa}  to report the performance of Charades-STA\cite{gao2017tall} and ActivityNet-Captions \cite{krishna2017dense} under R@1.

Novel-word OOD testing results are shown in Table~\ref{table:novel word}.  We collect the performance reported by \citet{visa} and  reproduce the baseline model \cite{mmn} with CLIP features under the same settings.   One can see that we outperform the SOTA method by a significant margin. Especially on Charades-STA dataset, we outperform the baseline model MMN \cite{mmn} with a margin of 2.62\%/4.46\% under the IoU = 0.5/0.7. We outperform VISA \cite{visa} by a margin of 4.12\%/7.75\%. One can see that on ActivityNet-Captions, the margin is less obvious than Charades-STA (2.21\% vs 4.12\%), it is partially because ActivityNet-Captions display longer moments than Charades-STA (averaged 37.14s vs 8.09s), so it is more challenging to capture the video change. Also, compared with the 37 long moments  ($L_{mom}$/$L_{vid}$ $\geq$ 0.5) in Charades-STA, there are over 15k in ActivityNet-Captions, which makes it trivial by predicting long predictions instead of capturing the semantics.

 Obtaining superior OOD performance over video-based~(Kinetics\cite{kay2017kinetics}) pre-training models demonstrates that our method can take advantage of the image-text pre-training feature and obtain generalisable correlations to unseen words and scenes.

\subsubsection{Original Split Testing }

To further evaluate the effectiveness of our method, we also conduct extensive experiments on the standard testing, where the training and testing splits share independent and identical distribution. 

In Table \ref{table:charades}, we include the recently reported results of SOTA methods as well as their visual pre-training data~(Vis.P.T). One can see that when replacing the MMN from separated pre-training with joint pre-training features, the performance increases from 47.31\% to 50.48\%, indicating that the correlation between the vision and text is essential. With our proposed method to empower the model's understanding of actions, the performance further improves to 52.32\%, outperforming the baseline model with a large margin of  5.01\%. 
From the results of ActivityNet-Captions, we can see that even though there is a performance drop from 
video-based pre-training to image-based pre-training (48.59\% vs 46.89\%), our method can fill the gap by injecting the video change understanding into the model.


To evaluate the learning of correlations, we compare with existing methods with a specific focus on image-based pre-training datasets\cite{deng2009imagenet}. As one can see from Table \ref{table:charades}, not only can we outperform SOTAs by a large margin, and we can narrow down the gap between the image-based pre-training feature and video-based pre-training feature.

\begin{table*}
\renewcommand\arraystretch{1.05}

  \centering
  \scalebox{0.90}{
  \begin{tabular}{l|c|cccc|c|cccc}
    \hline
     \multirow{3}{*}{Method} & \multicolumn{5}{c|}{Charades-STA} &\multicolumn{5}{c}{ ActivityNet-Captions}\\  
     \cline{2-11}
     &\multirow{2}{*}{\small Vis.P.T} & R@1,& R@1,& R@5,& R@5,&\multirow{2}{*}{ \small Vis.P.T}& R@1,& R@1,& R@5,& R@5,\\
           &&IoU=0.5&IoU=0.7&IoU=0.5&IoU=0.7 &&IoU=0.5&IoU=0.7&IoU=0.5&IoU=0.7 \\
    \hline
  
VideoBert \cite{sun2019videobert}   & \multirow{3}{*}{Video-Text} &32.70& 19.50& 68.10& 46.20 &\multirow{3}{*}{Video-Text} & 37.20& 21.00 & 66.70& 53.60 \\
MIL-NCE \cite{miech2020end} & & 37.00 & 21.20 & 74.30 & 50.40 & & 41.80 & 24.50& 73.50 & 57.70  \\
LocVTP \cite{cao2022locvtp}   &  & 43.60 & 26.30&81.90&55.30 && 48.20 & 30.50 & 80.10 & 64.70 \\
             \hline
   MGSL \cite{mgsl}  &  \multirow{2}{*}{Video}& 63.98 & 41.03 & 93.21& 63.85 &\multirow{8}{*}{Video}&51.87&31.42&82.60&66.71 \\
    D-TSG \cite{dtsg}  && 65.05 &42.77 & 94.42& 65.16&&54.29&33.64&86.58&69.36  \\

\cline{2-6}
       2D-TAN \cite{2dtan}  &\multirow{6}{*}{Image} &39.70 & 23.31 & 80.32 & 51.26&&44.51&26.54&77.13&61.96 \\
       VSL-Net \cite{vslnet}  &&39.20&20.80 & -&- & & 43.22&26.16 & -& -\\ 
    CBLN \cite{cbln} && 43.67  &24.44  & \textbf{88.39} & 56.49& & 48.12&27.60&79.32&63.41 \\
    DCM \cite{dcm}  &  & 47.80& 28.00 & - &- & & 44.90 & 27.70 & - &-\\
   DRN \cite{DRN}  &  & 42.90& 23.68 &87.80 &54.87 & & 45.45 & 24.36 & 77.97 & 50.30\\
    MMN \cite{mmn}  & &47.31&27.28&83.74&58.41 & & 48.59 & 29.26 & 79.50 & 64.76\\
         \hline

MMN \cite{mmn}  &\multirow{2}{*}{Image-Text} & 50.48 & 29.65& 85.27&60.67 &\multirow{2}{*}{Image-Text} &46.89 & 27.26 & 78.32 & 63.47\\
   \abbrname(Ours) & & \textbf{52.32} & \textbf{31.37}&87.03&\textbf{62.30} & & \textbf{48.09} & \textbf{28.76} & \textbf{79.69} & \textbf{64.88}  \\
    \hline
    
  \end{tabular}
  }
  \caption{ Comparisons to the state-of-the-art methods on the standard splits of VMR benchmark datasets.
  The ``Vis.P.T'' column indicates that the
  video encoders adopted are pre-trained by 
  only videos~\cite{kay2017kinetics} (``Video''), 
  only images~\cite{deng2009imagenet} (``Image''), 
  video-text pairs~\cite{miech2019howto100m} (``Video-Text'') and 
  Image-text pairs~\cite{clip} (``Image-Text''). Best performances among image-based pre-training methods are highlighted in bold.}
  \label{table:charades}
\end{table*}

\begin{figure*}
  \centering
  \includegraphics[width=2\columnwidth]{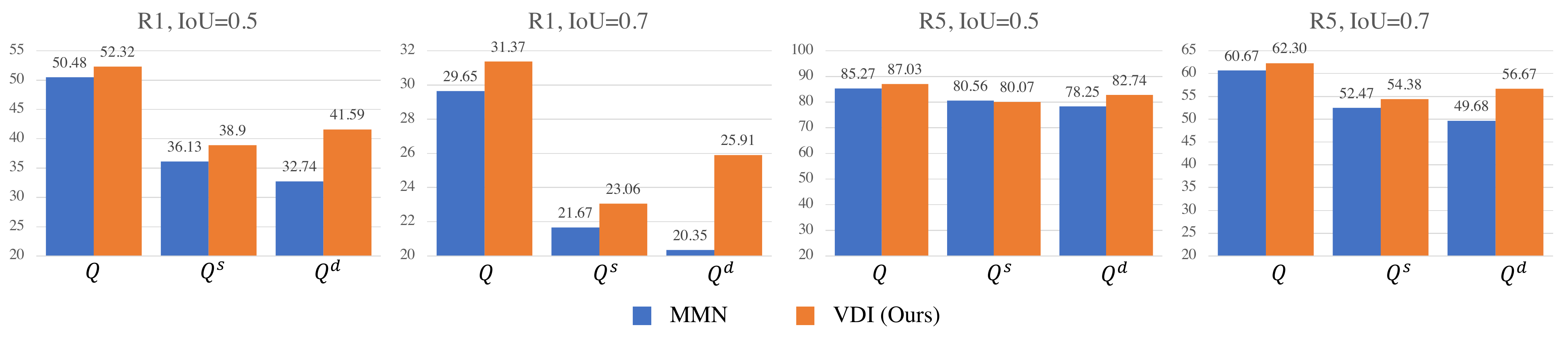}
  \caption{
  Video Moment Retrieval by the complete $Q$, 
  the static $Q^s$ or the dynamic $Q^d$ sentence descriptions.
  }
    \vspace{-1em}

  \label{fig:quan}
\end{figure*}

\subsubsection{Ablation Study}

In this section, we perform in-depth ablations to evaluate the effectiveness of each component in \abbrname on Charades-STA dataset \cite{gao2017tall} with novel-word splits \cite{visa}.  
We report the performance under R@1 for IoU $\in \{0.5,0.7\}$ and mIoU. 

\paragraph{Visual Context  Injection.} To evaluate that it is essential to inject visual context information into the text embeddings, we study different types of  visual context generations as shown in Table~\ref{table:ablation model}.  
We take the MMN with CLIP pre-training features as our baseline.
The superior performances yielded by the models with visual context injection over the baseline 
demonstrate the effectiveness of the design.
Moreover, 
we observe that
using static query to probe the videos ($\mathcal{L}^{vc}$ w/ $Q^s$) 
produces better results
than probing by the complete sentence ($\mathcal{L}^{vc}$ w/ $Q$)
and the pure visual context generation ($\mathcal{L}^{vc}$ w/o $Q$) without text.
This implies the importance to avoid correlating
the text descriptions of video changes with irrelevant visual context.
%


\paragraph{Spatial Dynamic Injection.} To evaluate that spatial dynamic information is essential for the text encoder, we study two types of dynamic modelling, including LSTM \cite{lstm} and Transformer Encoder \cite{vaswani2017attention}. As shown in Table \ref{table:ablation model},  by introducing spatial dynamic information, 
both a recurrent network ($\mathcal{L}^{sd}$ w/ LSTM)
or a transformer encoder ($\mathcal{L}^{sd}$ w/ Trans)
is ready to contribute to a more precise VMR.

With a combination of  $\mathcal{L}^{vc}$ and $\mathcal{L}^{sd}$, the performance improves to 46.47\%, which further demonstrates the effectiveness of \abbrname.

 \begin{table}
 \renewcommand\arraystretch{1.05}

  \setlength{\tabcolsep}{0.38cm}
  \centering
  \begin{tabular}{l|ccc}
    \hline
     \multirow{2}{*}{Method} 
     & R@1,& R@1, & \multirow{2}{*}{mIoU}\\
     &IoU=0.5&IoU=0.7& \\
    \hline
     Baseline 
     & 43.85 & 24.17 & 39.50 \\
     \hline
     $\mathcal{L}^{vc}$ w/o $Q$  &43.12 & 25.32 & 39.43\\
     $\mathcal{L}^{vc}$ w/ $Q$ &45.02 & 27.63 & 40.40\\
     $\mathcal{L}^{vc}$ w/ $Q^s$ &45.47&\textbf{29.35} & 40.61 \\
     \hline

     $\mathcal{L}^{sd}$ w/ LSTM & 45.32 & 25.76 & 40.07\\
     $\mathcal{L}^{sd}$ w/ Trans &44.60 & 26.06 & 40.09\\
     \hline
     $\mathcal{L}^{vc}+\mathcal{L}^{sd}$ & \textbf{46.47} &28.63 & \textbf{41.60} \\
        \hline
  \end{tabular}
  \caption{Ablation studies on visual context injection $\mathcal{L}^{vc}$ with different text queries and spatial dynamic injection $\mathcal{L}^{sd}$ with different sequence analysis models. }
  \label{table:ablation model}

\end{table}

\paragraph{Dynamics Awareness.}
We further evaluate our baseline and \abbrname models
by using either the complete sentences $Q$, the static $Q^s$ or the dynamic $Q^d$ queries to retrieve video moments on Charades-STA.
As shown in Fig.~\ref{fig:quan},
the baseline MMN model
yields better results when retrieving video 
moments
by static queries than the dynamic ones
while our \abbrname model is in opposed.
This verifies the sensitivity of \abbrname to video changes by correlating visual context and spatial dynamics with text
and explains its improvements to the baseline on all VMR tasks.

\section{Conclusion}

In this paper, we propose to learn universal visual-textual correlations
for video moment retrieval (VMR). To address the limitation that the image-text pre-training methods are less sensitive to video changes, we design visual context and spatial dynamic injection to the text encoder with an emphasis on the words describing video changes. By doing so, the potentially relevant
visual and motion patterns in videos are encoded in the corresponding text embeddings, enabling more
accurate video-text alignments.  Experiments on two important datasets (Charades-STA and ActivityNet-Captions) prove that \abbrname can learn both effective and generic visual-text correlations. Moreover, the comparison between the before and after visual-dynamic injection demonstrate the sensitivity of
\abbrname to video changes.

\section*{Acknowledgements}
This work was supported by the China Scholarship Council, the Alan Turing Institute Turing Fellowship, Veritone, Adobe Research and Zhejiang Lab (NO. 2022NB0AB05).

{\small
\bibliographystyle{plainnat}

\bibliography{egbib}
}

\end{document}